\documentclass[11pt,a4paper]{article}

\usepackage{graphicx}
\usepackage[export]{adjustbox}
\usepackage{arabtex}
\usepackage{utf8}
\usepackage[utf8x]{inputenc}
\usepackage{romannum}
\usepackage{array}
\usepackage{multicol}
\usepackage{comment}
\usepackage{caption}
\usepackage{subcaption}

\usepackage[hyperref]{acl2021}
\usepackage{times}
\usepackage{latexsym}

\usepackage{microtype}

\aclfinalcopy 

\title{LU-BZU at SemEval-2021 Task 2: \\
Word2Vec and Lemma2Vec performance in Arabic Word-in-Context disambiguation
}

\author{Moustafa Al-Hajj \\
Lebanese University \\
Lebanon \\
\texttt{moustafa.alhajj@ul.edu.lb} \\\And
Mustafa Jarrar \\
Birzeit University \\
Palestine \\
\texttt{mjarrar@birzeit.edu} \\}

\date{}

\begin{document}
\setcode{utf8}
\maketitle
\begin{abstract}
This paper presents a set of experiments to evaluate and compare between the performance of using CBOW Word2Vec and Lemma2Vec models 
for Arabic Word-in-Context (WiC) disambiguation  without using sense inventories or sense embeddings. As part of the SemEval-2021 Shared Task 2 on WiC disambiguation, we used the \texttt{\small dev.ar-ar} dataset (2k sentence pairs) to decide whether two words in a given sentence pair carry the same meaning. 
We used two Word2Vec models: Wiki-CBOW, a pre-trained model on Arabic Wikipedia, and another model we trained on large Arabic corpora of about 3 billion tokens. Two Lemma2Vec models was also constructed based on the two Word2Vec models. Each of the four models was then used in the WiC disambiguation task, and then evaluated on the SemEval-2021 \texttt{\small test.ar-ar} dataset. At the end, we reported the performance of different models and compared between using lemma-based and word-based models. 

\end{abstract}

\section{Introduction}
\label{ssec:introduction}
As a word may denote multiple meanings ({\em i.e.}, senses) in different contexts, disambiguating them is important for many NLP applications, such as information retrieval, machine translation, summarization, among others. For example, the word “table” in sentences like “I am cleaning the table”, “I am serving the table”, “I am emailing the table”, refer to “furniture”, “people”, and “data” respectively. Disambiguating the sense that a word denotes in a given sentence is important for understanding the semantics of this sentence. 

To automatically disambiguate word senses in a given context, many approaches have been proposed based on supervised, semi-supervised, or unsupervised learning models. Supervised and semi-supervised methods rely on full, or partial, labeling of the word senses in the training corpus to construct a model \citep{lee2002empirical,klein2002combining}. On the other hand, unsupervised approaches induce senses from unannotated raw corpora and do not use lexical resources. The problem in such approaches, is that unsupervised learning of word embeddings produces a single vector for each word in all contexts, and thus ignoring its polysemy. Such approaches are called static Word Embeddings. To overcome the problem, two types of approaches are suggested \citep{pilehvar2018wic}: multi-prototype embeddings, and contextualized word embeddings. The latter suggests to model context embeddings as a dynamic contextualized word representation in order to represent complex characteristics of word use. 
Proposed architectures such as ELMo \citep{peters2018deep}, ULMFiT \citep{DBLP:journals/corr/abs-1801-06146}, GPT \citep{radford2018improving}, T5 \citep{DBLP:journals/corr/abs-1910-10683}, and BERT \citep{DBLP:journals/corr/abs-1810-04805}, achieved breakthrough performance on a wide range of natural language processing tasks. 
In multi-prototype embeddings, a set of embedding vectors are computed for each word, representing its senses. In \citep{pelevina2017making}, multi-prototype embeddings are produced based on the embeddings of a word. As such, a graph of similar words is constructed, then similar words are grouped into multiple clusters, each cluster representing a sense. As for \citet{mancini2016embedding}, multi-prototype embeddings are produced by learning word and sense embeddings jointly from both, a corpus and a semantic network. In this paper we aim at using static word embeddings for WiC disambiguation.

Works on Arabic Word Sense Disambiguation (WSD) are limited, and the proposed approaches are lacking a decent or common evaluation framework. Additionally, there are some specificities of the Arabic language that might not be known in other languages. Although polysemy and disambiguating are challenging issues in all languages, they might be more challenging in the case of Arabic \citep{jarrar2018diacritic,jarrar2019arabic} and this for many reasons. For example, the word šāhd ({\scriptsize \<شاهد>}) could be šāhid ({\scriptsize \<شاهِد>}) which means a \textit{witness}, or šāhada ({\scriptsize \<شاهَدَ>}) which means \textit{watch}. As such, disambiguating words senses in Arabic, is similar to disambiguating senses of English words written without vowels. Second, Arabic is a highly inflected and derivational language. As such, thousands of different word forms could be inflected and derived from the same stem. Therefore, words in word embeddings models will be considered as different, 
which may affect the accuracy and the utility of their representation vectors, as the same meaning could be incarnated in distributed word forms in corpora, which has led some researchers to think that using lemma-based models might be better than using word-based embeddings in Arabic \citep{salama2018, shapiro2018}. This idea will be discussed later in sections \ref{ssec:lemma2vec} and \ref{ssec:experiments}.

\citet{alkhatlan2018word} suggested an Arabic WSD approach based on Stem2Vec and Sense2Vec. The Stem2Vec is produced by training word embeddings after stemming a corpus, whereas the Sense2Vec is produced based on the Arabic WordNet sense inventory, such that each synset is represented by a vector. To determine the sense of a given word in a sentence, the sentence vector is compared with every sense vector, then the sense with maximum similarity is selected.

\citet{laatar2017word} did not use either stemming or lemmatization. Instead, they proposed to determine the sense of a word in context by comparing the context vector with a set of sense vectors, then the vector with the maximum similarity is selected. The context vector is computed as the sum of vectors of all words in a given context, which are learnt from a corpus of historical Arabic. On the other hand, sense vectors are produced based on dictionary glosses. Each sense vector is computed as the sum of vectors (learnt from the historical Arabic corpus) of all words in the gloss of a word.

Other approaches to Arabic WSD \citep{elayeb2019arabic} employ other techniques in machine learning and knowledge-based methods \citep{bouhriz2016word,bousmaha2013approche,soudani2014generic,merhbene2014approche,al2015extract,bounhas2015experimenting}.

In this paper, we present a set of experiments to evaluate the performance of using Lemma2Vec vs CBOW Word2Vec in Arabic WiC disambiguation. 
The paper is structured as follows: Section~\ref{ssec:background} presetns the background of this work. Section~\ref{ssec:overview} overviews the WiC disambiguation system. Section~\ref{ssec:corpus} and Section~\ref{ssec:lemma2vec}, respectively, present the Word2Vec and Lemma2Vec models. In Section~\ref{ssec:experiments} we present the experiments and the results; and in section~\ref{ssec:conclusions} we summarize our conclusions and future work.

\section{Background}
\label{ssec:background}
Experiments presented in this paper are part of the SemEval shared task for Word-in-Context disambiguation \citep{martelli-etal-2021-mclwic}.

\begin{figure}[h]
    \centering
    \includegraphics[width=0.5\textwidth, frame]{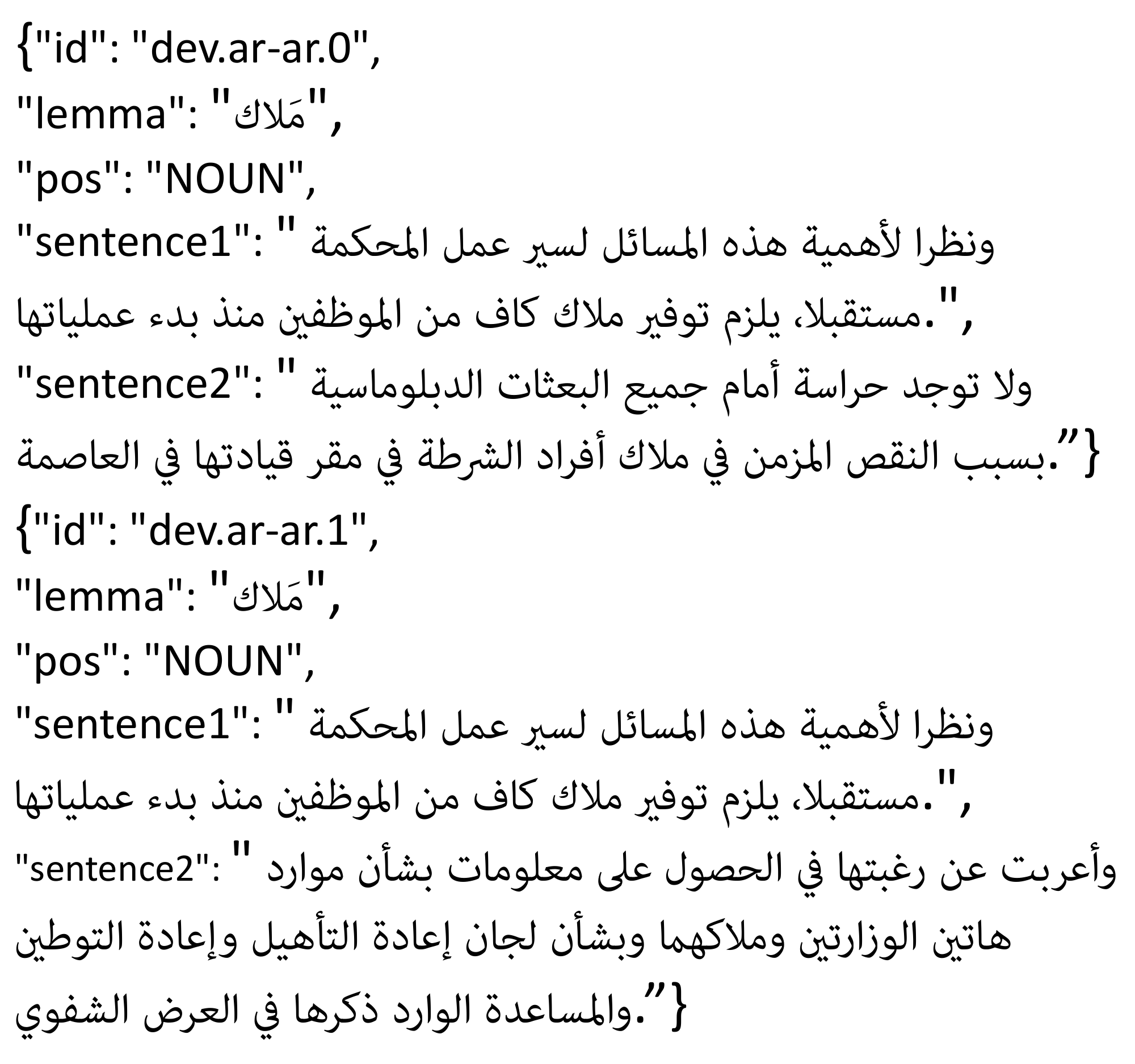}\par
    \caption{Two examples of sentence pairs.}
    \label{fig:pairsentences}
\end{figure}

The task aims at capturing the polysemous nature of words without relying on a fixed sense inventory.
A common evaluation dataset is provided to participants in five languages, including Arabic, our target language in this paper.
The dataset was carefully designed to include all parts of speeches and to cover many domains and genres. 
The Arabic dataset (called multilingual \texttt{\small ar-ar}) consists of two sets: a train set of 1000 sentence pairs for which tags (TRUE or FALSE) are revealed, and a test set of 1000 sentence pairs for which tags were kept hidden during the competition. 
Figure~\ref{fig:pairsentences} gives two examples of sentence pairs in the \texttt{\small dev.ar-ar} dataset. 
Each sentence pair has a word in common for which start and end positions in sentences are provided. Participants in the shared task were asked to infer whether the target word carries the same meaning (TRUE) or not (FALSE) in the two sentences.

\section{System Overview}
\label{ssec:overview}
This section describes our method to Arabic WiC disambiguation based on two types of embeddings: CBOW Word2Vec and Lemma2Vec.

\begin{figure}[ht]
    \centering
    \includegraphics[trim=0 101 0 87,clip,width=\linewidth, frame]{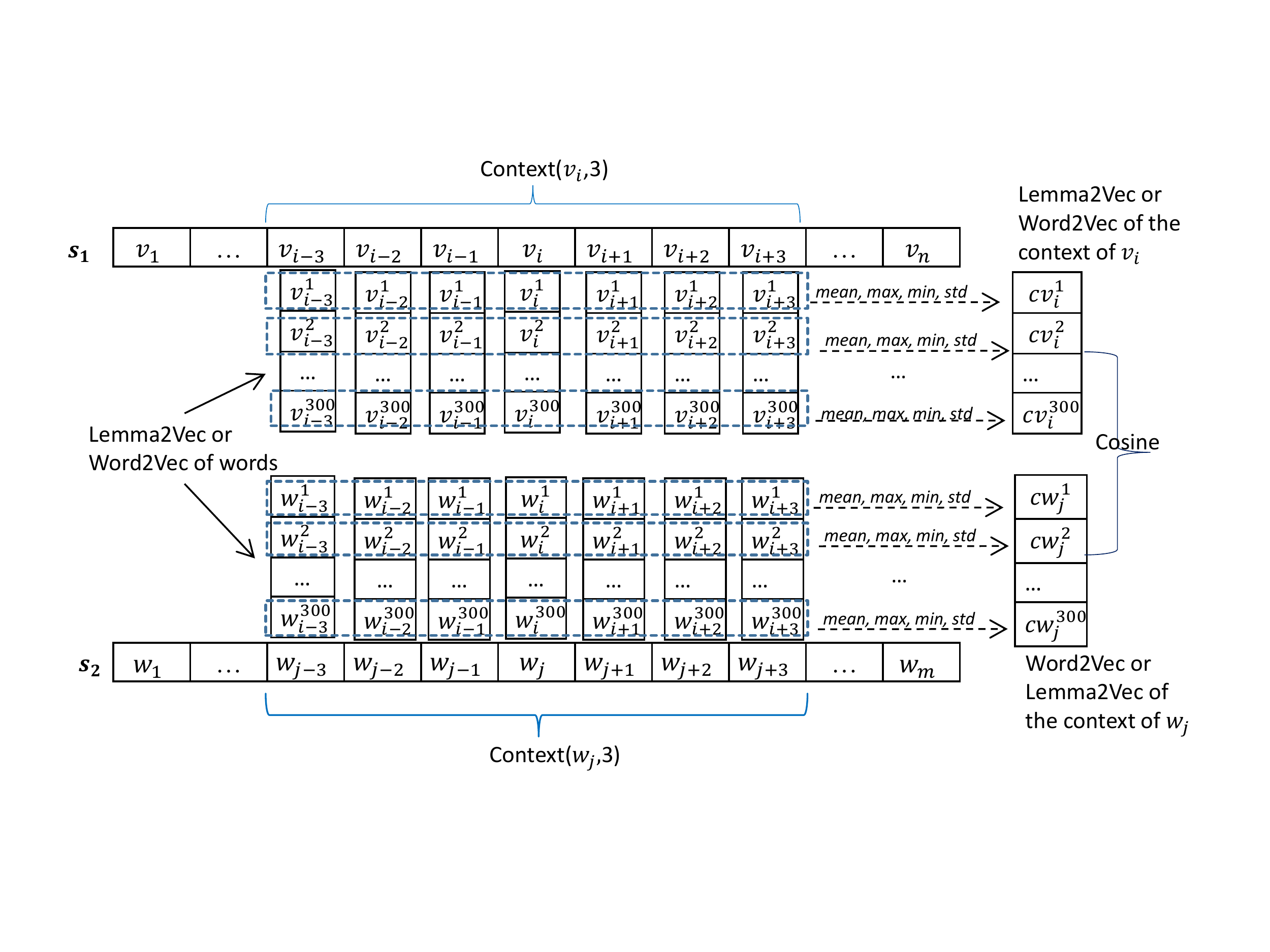}\par
    \caption{Calculation of $context(v_i,3)$ and $context(w_j,3)$ vectors and the cosine similarity between.}
    \label{fig:cosine}
\end{figure}

Given two sentences, $s_1$ and $s_2$ , and two words, $v_i$  from $s_1$  and $w_j$  from $s_2$ , the objective is to check whether $v_i$  and $w_j$  have the same meaning. To this end, our system extracts contexts of $v_i$  and $w_j$  from the sentence pair, represents them in two vectors and finally compares the two resulting vectors using the cosine similarity.
The context of a word $w$ of size $n$ (denoted by  $context(w,n)$) is composed of the words that surround the word $w$, with $n$ words on the left and $n$ words on the right ($n$ varying between $1$ and $10$ in conducted experiments). To represent $context(w,n)$ in a vector space, two methods are proposed: first one is based on CBOW Word2Vec embeddings vectors \citep{mikolov2013efficient} of the words appearing in the context, whereas the second is based on the Lemma2Vec of lemmas of words appearing in the context. To select the best way to represent the $context(w,n)$ by a vector, classification experiments were conducted using 
(\romannum{1}) different pooling operations, $min$, $max$, $mean$, and $std$ to combine words/lemmas vectors of the context, 
(\romannum{2}) different threshold values (between 0.55 and 0.85) and 
(\romannum{3}) the removal of functional words (also called stop words). The later are used to express grammatical relationships among other words, they are characterized by they high frequency in the corpus which might affect the WiC disambiguation accuracy. 
The cosine similarity is then used to compare vectors of $context(v_i,n)$ and $context(w_j,n)$. Figure \ref{fig:cosine} illustrates how the cosine similarity is calculated from $context(v_i,3)$ and $context(w_j,3)$.   

Classification experiments on SemEval-2021 \texttt{\small ar-ar} datasets were conducted using the following two CBOW Word2Vec models and two corresponding Lemma2Vec models: 
(\romannum{1}) Wiki-CBOW, a pre-trained Word2Vec model from the set of AraVec models \citep{soliman2017aravec} , 
(\romannum{2}) our CBOW Word2Vec model that we trained ourselves, 
(\romannum{3}) Lemma2Vec model that we constructed based on the Wiki-CBOW model, and 
(\romannum{4}) Lemma2Vec that we constructed based on our CBOW Word2Vec model. 
Based on these four models, four experiments were conducted to tune the following parameters: context size ($context\_size$), $threshold$, pooling operation ($pooling$) and removing of functional words ($stop\_words$). 

\section{Corpus and trained Word Embeddings}
\label{ssec:corpus}
Two CBOW Word2Vec models were used in our experiments. The Wiki-CBOW \citep{soliman2017aravec}, which consists of 234,173 vocabulary size, and another model we trained our self which consists of 334,161 vocabulary size. The Wiki-CBOW model was learnt from a corpus of Arabic Wikipedia articles of about 78 million words,  the principal hyper-parameters are: 5 for minimum word count and 5 for window size.

Our CBOW Word2Vec model was trained on Modern Standard Arabic corpora, such as \citep{elabu,abbas2005comparison,abdelali2014amara} of about 3 billion words; it was fit using 300-dimensional word vectors, 100 the minimum count of words, training epochs of 5 and window size of 5.

Before training the Word2Vec model, several normalization and preprocessing steps were performed. First, all diacritics, punctuations, Madda character, digits (Hindi and Arabic), Latin characters (including accented letters) were removed. Second, some special Arabic letters are unified. Third, sequences of repeated characters with length larger than 2 were reduced to one character; repeated spaces were also replaced by one space. Fourth, different forms of Alifs ({\scriptsize \<أ إ آ>}) are replaced with ({\scriptsize \<ا>}). Spaces followed by a period character and new lines were considered to be end of sentence marks. The split method in Python is used in tokenization. The vocabulary size of the resulted model is 334,161.

\section{Constructing the Lemma2Vec models}
\label{ssec:lemma2vec}
Two Lemma2Vec models were produced, based on both: the Wiki-CBOW Word2Vec model, and our CBOW Word2Vec  model. 
Each vocabulary in each of the Word2Vec models was lemmatized first. 
Then a vector for each lemma ( {\em i.e.}, Lemma2Vec) is calculated as following: 
first all word forms belonging to this lemma are fetched, 
then their Word2Vec vectors are combined through a $mean$ pooling operation.
The lemmatization process was performed using in-house tools and lexicographic databases 
\footnote{\url{https://ontology.birzeit.edu}} belonging to Birzeit University \citep{jarrar2019arabic,jarrar2019arabic1, jarrar2019lemon}. 
In case of a word cannot be lemmatized due to misspelling, incorrect tokenization or in case of foreign word (not included in our database), then the corresponding Lemma2Vec is considered to be its Word2Vec vector.

Table~\ref{table:lemmatization} summarizes the lemmatization results that we performed on both, the Wiki-CBOW model and our CBOW Word2Vec model. The lemmatized words of SemEval-2021 all \texttt{\small ar-ar} dataset, as well as the Word2Vec and Lemma2Vec of \texttt{\small ar-ar} datasets words’s vectors used in this paper are available on-line~\footnote{\url{https://ontology.birzeit.edu/semeval2021_data.zip}}.

\begin{table}[ht]
\centering
\begin{tabular}{|@{ }m{7.2em}|@{}>{\centering\arraybackslash}m{4.9em}@{}|@{}>{\centering\arraybackslash}m{5.7em}@{}|}
\hline & {\small \textbf{Wiki-CBOW}} & {\small \textbf{Our Word2Vec}} \\ 
& {\small 78M words\newline min\_count 5} & {\small 3B words \newline min\_count 100} \\ \hline
\hline
{\small Unique word forms} & {\small 234,173} & {\small 334,161} \\ \hline
{\small Unique lemmas} & {\small 100},{\small 040} & {\small 54,788} \\ \hline
{\small Words not lemmatized} & {\small 22,054} & {\small 28,098}\\
\hline
\end{tabular}
\caption{\label{table:lemmatization} Lemmatization results for both models. }
\end{table}

\section{Experiments Results and Discussion}
\label{ssec:experiments}
Given our Arabic WiC disambigation method described in Section~\ref{ssec:overview}, 
and given the SemEval multilingual \texttt{\small dev.ar-ar} dataset provided by SemEval-2021 \citep{martelli-etal-2021-mclwic}, 
four classification experiments were conducted using the cosine similarity and based on the two Word2Vec models and the two Lemma2Vec models.
The objective is to tune the following parameters for each model:  $context\_size$ (ranging from 1 to 10), $threshold$ (we determined empirically the range from 0.55 to 0.85 with 0.1 step size), $pooling$ ($min$, $max$, $mean$ and $std$), and $stop\_words$ ($yes$, $no$). 
Then the values of parameters corresponding to the high F1-scores for TRUE (T) and FALSE (F) classes are selected in order to classify sentence pairs in the \texttt{\small test.ar-ar} dataset. For each model we did the following to find the high F1-scores for T and F: 
For each $context\_size$ (between 1 and 10) and for each value of the $stop\_words$ ($yes$ or $no$) we plotted 8 line plots (4 for T and 4 for F) for each of the four pooling operations ($mean$, $max$, $min$ and $std$) and for $threshold$ ranging from 0.55 to 0.85 ({\em i.e.}, 20 plots for each model, resulting 80 plots). 

\begin{figure*}[t]
\begin{multicols}{2}
\begin{subfigure}[b]{1\linewidth}
    \centering
    \includegraphics[trim=0 7.5 0 20,clip,width=\linewidth]{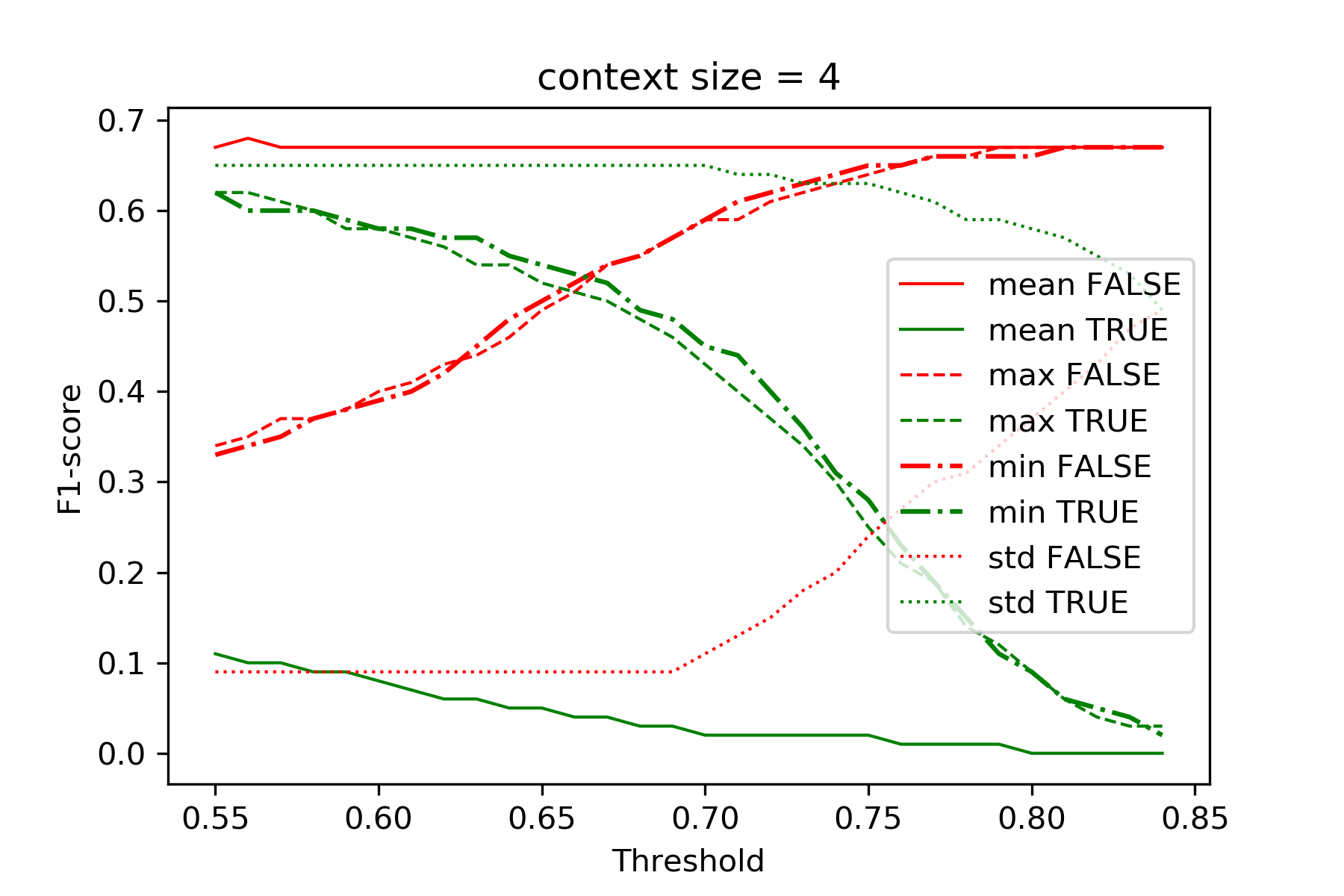}\par 
    \vspace{-0.3\baselineskip}
    \caption{\textbf{Wiki-CBOW Word2Vec} model. \\
$context\_size = 4$ \ - \ \  $pooling = min$\\
$threshold = 0.66$ \ - \ $stop\_words = yes$
}
    \label{fig:f1scores1}
\end{subfigure}

\begin{subfigure}[b]{1\linewidth}
    \centering
    \includegraphics[trim=0 7.5 0 20,clip,width=\linewidth]{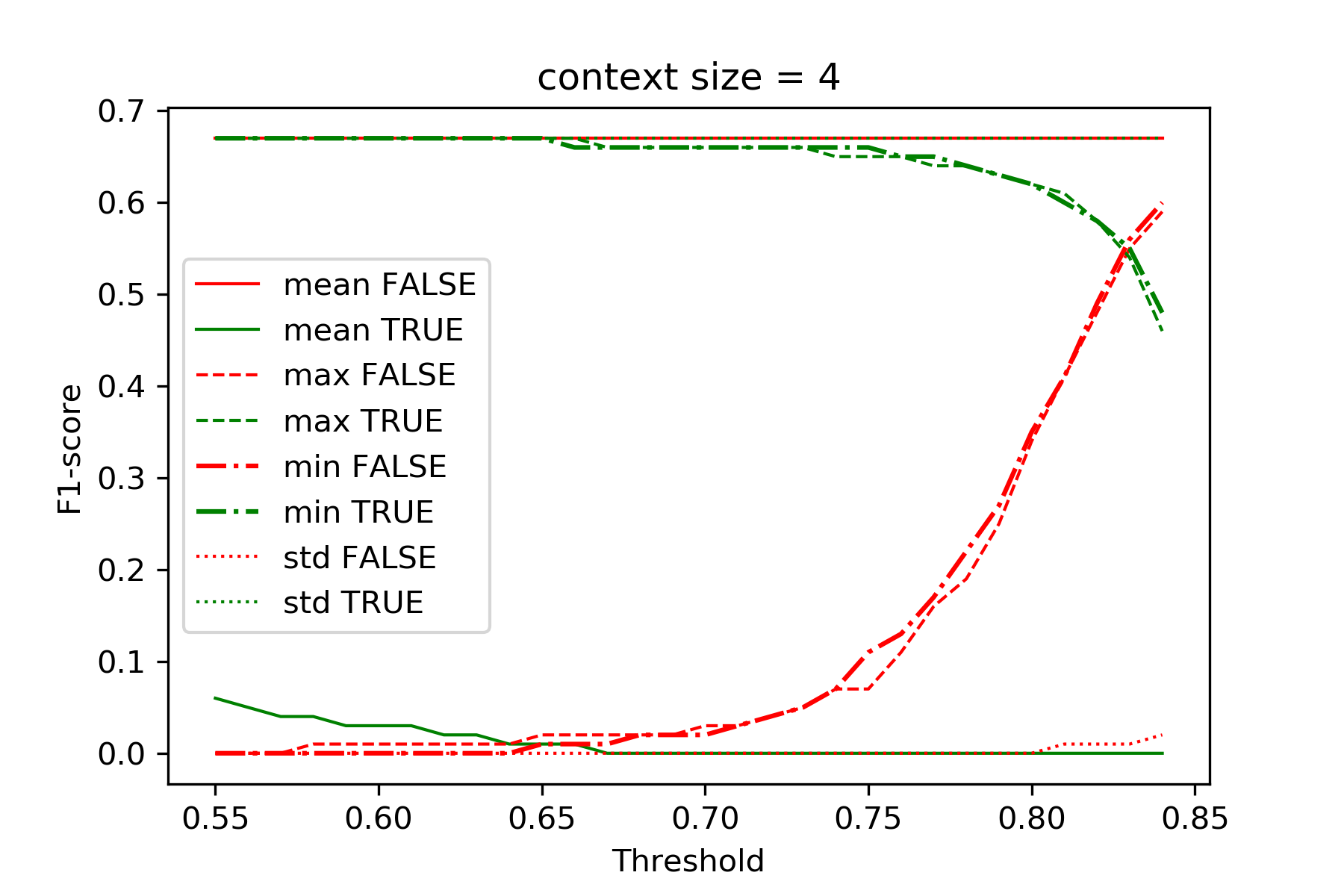}\par 
    \vspace{-0.3\baselineskip}
    \caption{\textbf{our Word2Vec} model. \\
$context\_size = 4$ \ - \ \ $pooling = min$\\
$threshold = 0.83$  \ - \ $stop\_words = no$}
    \label{fig:f1scores2}
\end{subfigure}
\end{multicols}
\vspace{-1.5\baselineskip}
\begin{multicols}{2}

\begin{subfigure}[b]{1\linewidth}
    \centering
    \includegraphics[trim=0 7.5 0 20,clip,width=\linewidth]{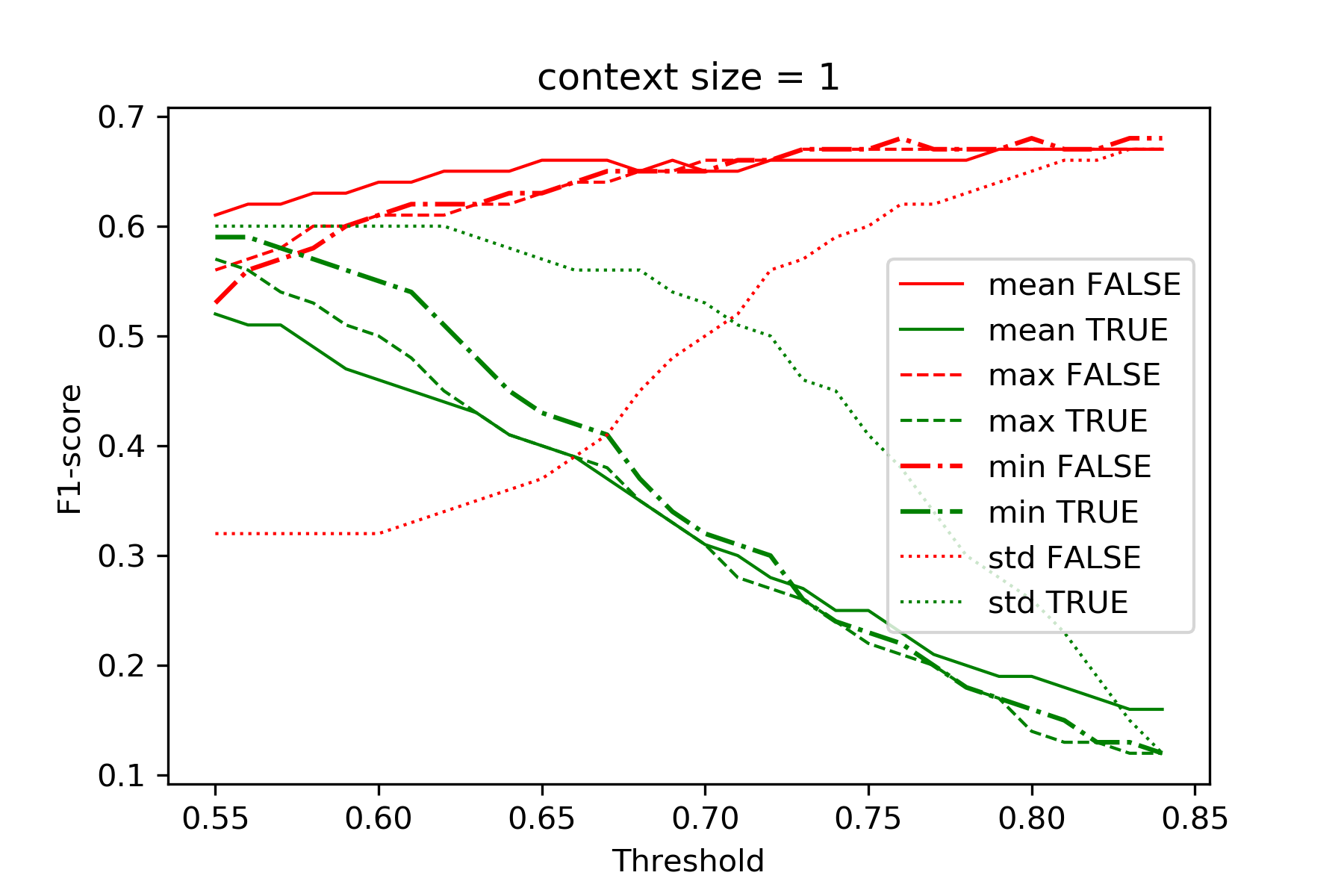}\par 
    \vspace{-0.3\baselineskip}
    \caption{\textbf{Wiki-CBOW Lemma2Vec} model. \\
$context\_size = 1$ \ - \ \ $pooling = min$\\
$threshold = 0.56$  \ - \ $stop\_words = yes$}
    \label{fig:f1scores3}
\end{subfigure}

\begin{subfigure}[b]{1\linewidth}
    \centering
    \includegraphics[trim=0 7.5 0 20,clip,width=\linewidth]{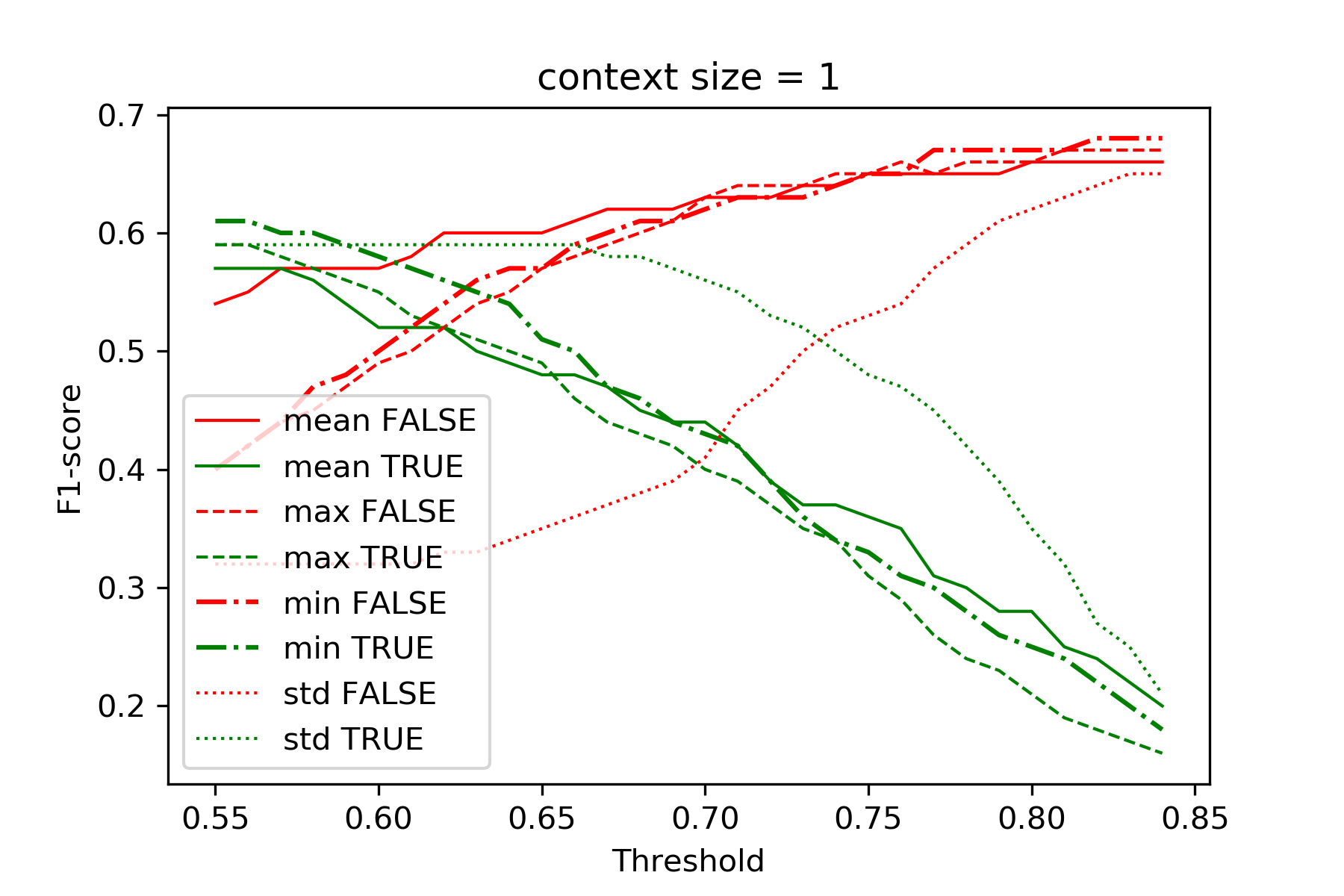}\par 
    \vspace{-0.3\baselineskip}
    \caption{\textbf{our Lemma2Vec} model. \\
$context\_size = 1$ \ - \ \ $pooling = mean$\\
$threshold = 0.58$  \ - \ $stop\_words = yes$}
    \label{fig:f1scores4}
\end{subfigure}
\end{multicols}
\vspace{-1.4\baselineskip}
\caption{The best four F1-scores markers plots for each of the four models.The values of parameters are under each plot.}
\label{fig:f1scores}
\end{figure*}

Figures \ref{fig:f1scores1}, \ref{fig:f1scores2}, \ref{fig:f1scores3} and \ref{fig:f1scores4} show the best 4 F1-scores line plots for each of the four models, and Table \ref{table:res1} shows the effective F1-scores values for T and F classes as well as precision and recall values (best results marked in bold). The values of parameters corresponding to the best result were then used in classifying the \texttt{\small test.ar-ar} dataset. 
The accuracies are reported in Table~\ref{table:res1} as well.

\begin{table}[t!]
\centering
\begin{tabular}{|@{ }m{3.4em}|>{\centering\arraybackslash}m{3.1em}|>{\centering\arraybackslash}m{3.5em}|>{\centering\arraybackslash}m{3.1em}|>{\centering\arraybackslash}m{3.5em}|}
\hline & \textbf{Exp1} & \textbf{Exp2} & \textbf{Exp3} &    \textbf{Exp4} \\  \hline \hline
\textbf{\small Model} &{\small Word2Vec Wiki-CBOW} &{\small Lemma2Vec  Wiki-CBOW} & {\small Word2Vec our model} & {\small Lemma2Vec our model} \\ \hline
{ \scriptsize $context\_size$} & 4 & 1 & 4 & 1\\ \hline
{\small $pooling$} & $min$ & $min$ & $min$ & $mean$\\ \hline
{\small $threshold$}& 0.66& 0.56& 0.83& 0.58\\ \hline
{\scriptsize $stop\_words$}& $yes$ & $yes$ & $no$ & $yes$ \\ \hline\hline

\textbf{Dataset}&\multicolumn{3}{c}{\texttt{\small dev.ar-ar}}&\\ \hline\hline

Tag & \begin{tabular}{ @{ }c| c }  T & F \end{tabular}&\begin{tabular}{@{ }c |c } T & F  \end{tabular} & \begin{tabular}{@{ }c|c} T & F \end{tabular} & \begin{tabular}{@{ }c|c} T &F\end{tabular} \\ \hline
\hline

{\small Precision} & \begin{tabular}{@{}c|c} 52 & 52 \end{tabular}&\begin{tabular}{@{}c |c } 57 & 58  \end{tabular} & \begin{tabular}{@{}c| c} 56 & 56 \end{tabular} & \begin{tabular}{@{} c| c } \textbf{56} & \textbf{56}  \end{tabular} \\ \hline

{\small Recall} & \begin{tabular}{ @{}c| c } 54 & 51 \end{tabular}&\begin{tabular}{@{}  c |c }  61 & 53  \end{tabular} & \begin{tabular}{@{}c |c} 55 &  57 \end{tabular} & \begin{tabular}{ @{}c |c } \textbf{55}  & \textbf{58}  \end{tabular} \\ \hline

{\small F1-score} & \begin{tabular}{ @{}c |c }  53  & 52 \end{tabular}&\begin{tabular}{ @{} c |c } 59   & 56 \end{tabular} & \begin{tabular}{@{}c| c} 55  &56 \end{tabular} & \begin{tabular}{ @{}c |c } \textbf{56} & \textbf{57}  \end{tabular} \\
\hline\hline

\textbf{Dataset}&\multicolumn{3}{c}{\texttt{\small test.ar-ar}}&\\ \hline\hline
{\small Accuracy} & 57 & 59 & 59 & \textbf{60} \\ \hline 

\end{tabular}
\caption{\label{table:res1} Best F1-score, precision and recall values of the four experiments on \texttt{\small dev.ar-ar} dataset with the values of tuned parameters. Below are accuracies  on \texttt{\small test.ar-ar} dataset.}
\end{table}
As shown in Figure \ref{fig:f1scores},  the Lemma2Vec models have the tendency to perform better 
with shorter context sizes compared with the Word2Vec models. A possible reason may be that, 
in case of Lemma2Vec, the narrow meaning of words is affected due to the increase number of words involved in Lemma2Vec vector calculation. 
The impact of Lemma2Vec on the narrow meaning of words is discussed in the next subsection.

The results with $yes$ for $stop\_words$ are slightly better but not significant. 
Additionally, the $min$ pooling was generally the best operation to combine the context vectors, and the results of both $min$ and $max$ pooling were close to each other.

\subsection{Lemma2Vec-Word2Vec Error Analyses}
\label{ssec:error}
This subsection discusses the performance of using lemma-based vs. word-based models in the WiC disambiguation task, which we summarize in Table~\ref{table:wiki_models} and Table~\ref{table:our_models}.  

\begin{table}[h]
\centering
\begin{tabular}{|m{9.5em}|@{ }>{\centering\arraybackslash}m{2em}|@{ }>{\centering\arraybackslash}m{2.5em}|>{\centering\arraybackslash}m{2em}|}
\hline & {TRUE} & {FALSE} & {Total}\\ 
\hline
{\small Correct L2V - Correct W2V} & 225 & 145 & 370\\ \hline
\textbf{\small Correct L2V - Wrong W2V} & 118 & 98 & 216 \\ \hline
\textbf{\small Wrong L2V - Correct W2V} & 66 & 116 & 182 \\
\hline
{\small Wrong L2V - Wrong W2V} & 91 & 141 & 232\\ \hline
\hline
Total & 500 & 500 & 1000\\ \hline
\end{tabular}
\caption{\label{table:wiki_models} Wiki-CBOW Lemma2Vec vs. Word2Vec }
\end{table}

Table~\ref{table:wiki_models} presents the results of experiments 1 and 2 (using Word2Vec and Lemma2Vec of Wiki-CBOW) 
whereas Table~\ref{table:our_models} presents the results of experiments 3 and 4 (using Word2Vec and Lemma2Vec of our CBOW model).  
In each table, we compare between cases that were correctly or wrongly classified by both models. 
For example, the second row in Table~\ref{table:wiki_models} shows that 216 sentence pairs (118 TRUE class + 98 FALSE class) were correctly classified using the Wiki-CBOW's Lemma2Vec model but wrongly classified using the Word2Vec. 
Similarly, 182 sentence pairs in the third row were correctly classified 
using the Word2Vec but wrongly classified 
using the Lemma2Vec.

\begin{table}[t]
\centering
\begin{tabular}{|m{9.5em}|@{ }>{\centering\arraybackslash}m{2em}|@{ }>{\centering\arraybackslash}m{2.5em}|>{\centering\arraybackslash}m{2em}|}
\hline & {TRUE} & {FALSE} & {Total}\\ 
\hline
{\small Correct L2V - Correct W2V} & 124 & 241 & 365\\ \hline
\textbf{\small Correct L2V - Wrong W2V} & 188 & 45 & 233 \\ \hline
\textbf{\small Wrong L2V - Correct W2V} & 58 & 178 & 236 \\
\hline
{\small Wrong L2V - Wrong W2V} & 130 & 36 & 166\\ \hline
\hline
Total & 500 & 500 & 1000\\ \hline
\end{tabular}
\caption{\label{table:our_models} Our Lemma2Vec vs. our Word2Vec }
\end{table}

As shown in both tables' second and third rows, the Lemma2Vec did not significantly improve the overall results; but notably, the Lemma2Vec shows a significant improvement over Word2Vec for TRUE class whereas Word2Vec is better for FALSE class. 

This conclusion is valid for all models, whatever are the corpora content, size and $min\_count$ hyperparameter. 

\begin{figure}[t]
    \centering
    \includegraphics[width=0.5\textwidth]{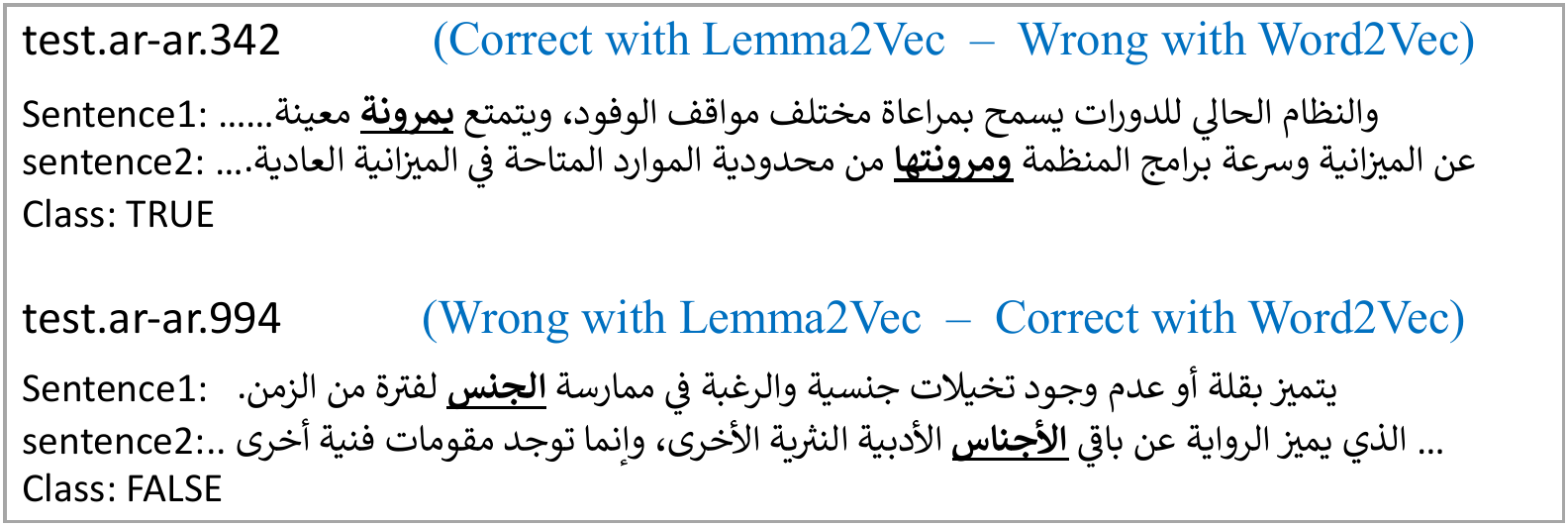}
    \caption{Example of errors.}
    \label{fig:example_of_errors}
\end{figure}

To understand the gain and loss by the lemma-based models, we manually analyzed most cases. Figure~\ref{fig:example_of_errors} illustrates such cases. The first sentence pair in Figure~\ref{fig:example_of_errors} was correctly classified by the Lemma2Vec (in \textbf{Exp4}) and wrongly by the Word2Vec (in \textbf{Exp3}). This illustrates that the lemma vector as a generalized model for its inflections ({\em i.e.}, a mean of word forms’ vectors) was better in deciding that both contexts are similar and that the two word forms have the same meaning. However, the second example in Figure~\ref{fig:example_of_errors} illustrates the other way. The Lemma2Vec was too general, and the Word2Vec was specific enough, to decide that the two word forms, in the two contexts, are different. The word from al-ǧins ({\scriptsize \<الجنس>}) could mean both \textit{genus} and \textit{sex}; however the other word form al-ʾaǧnās ({\scriptsize \<الأجناس>}), is semantically distinctive by its own morphology - as it can only be plural of \textit{genus}, and cannot be plural of \textit{sex}.

To conclude, although Lemma2Vec outperforms Word2Vec in some cases (mostly in the TRUE sentence pairs class), it underperforms Word2Vec in others cases (mostly in the FALSE sentence pairs class).  Since the distribution of TRUE and FALSE is equal in the datasets, 
the overall performance of both models is close to each other. Nevertheless, in case of an application scenario where a large proportion of sentence pairs is expected to be TRUE, we recommend the use of Lemma2Vec, otherwise the Word2Vec.

\section{Conclusions and Further Work}
\label{ssec:conclusions}
We presented a set of experiments to evaluate the performance of using Word2Vec and Lemma2Vec models in Arabic WiC disambiguation, without using external resources or any context/sense embedding model. Different models were constructed based on two different corpora, and different types of parameters were tuned. 
The final results demonstrated that Lemma2Vec models are slightly better than Word2Vec models for Arabic WiC disambiguation. More specifically, we found that Lemma2Vec outperforms Word2Vec for TRUE sentence pairs, but underperforms it for FALSE sentence pairs.  

We plan to extend our work to use our Lemma2Vec model to build a multi-prototype embeddings using the large lexicographic database available at Birzeit University. We plan also to fine tune the recently released Arabic BERT models, such as  \citep{safaya-etal-2020-kuisail,antoun2020arabert,abdelali2021pretraining,inoue-etal-2021-interplay}, using the same database. 

\section*{Acknowledgments}
We would like to thank the shared task organizers and the reviewers for their valuable comments and efforts towards improving our manuscript. We would like to also thank Taymaa Hammouda for her technical support. 

\bibliographystyle{acl_natbib}

\end{document}